\documentclass{article}

\usepackage{arxiv}

\usepackage[utf8]{inputenc} % allow utf-8 input
\usepackage[T1]{fontenc}    % use 8-bit T1 fonts
\usepackage{hyperref}       % hyperlinks
\usepackage{url}            % simple URL typesetting
\usepackage{booktabs}       % professional-quality tables
\usepackage{amsfonts}       % blackboard math symbols
\usepackage{nicefrac}       % compact symbols for 1/2, etc.
\usepackage{microtype}      % microtypography
\usepackage{lipsum}		% Can be removed after putting your text content
\usepackage{graphicx}
\graphicspath{ {./} }
\usepackage{natbib}
\usepackage{doi}
\usepackage{makecell}

\title{AlertTrap: A study on object detection in remote insect trap monitoring system using on-the-edge deep learning platform}

\date{March 4, 2022}	% Here you can change the date presented in the paper title
\author{\hspace{1mm}An D. ~Le \\
	Electrical and Computer Engineering Department\\
	University of California San Diego\\
	San Diego, USA \\
	\texttt{d0le@ucsd.edu} \\
	\And
	\hspace{1mm}Duy A. ~Pham \\
	Computer Science Department\\
	Bonn-Rhein-Sieg University of Applied Sciences\\
	Bonn, Germany \\
	\texttt{duyanhpham@outlook.com} \\
	\AND
	\hspace{1mm}Dong T. ~Pham \\
	Graduate School of Bioresource and Bioenvironmental Sciences\\
	Kyushu University\\
	Fukuoka, Japan \\
	\texttt{phamtdong0406@gmail.com} \\
	\And
	\hspace{1mm}Hien B. ~Vo \thanks{Correspondence author}  \\
	Electrical and Computer Engineering Department\\
	Vietnamese-German University\\
	Binh Duong, Vietnam \\
	\texttt{hien.vb@vgu.edu.vn} \\
}

\renewcommand{\headeright}{Article}
\renewcommand{\undertitle}{Article}
\renewcommand{\shorttitle}{AlertTrap}

\hypersetup{
pdftitle={AlertTrap: A study on object detection in remote insect trap monitoring system using on-the-edge deep learning platform},
pdfsubject={cs.CV},
pdfauthor={An D. ~Le, Duy A. ~Pham, Dong T. ~Pham, Hien B. ~Vo},
pdfkeywords={fruit fly, environmental data, smart IoT, edge computing, YOLOv4-tiny, MobileNet, Raspberry Pi, TPU},
}

\begin{document}
\maketitle

\begin{abstract}
	Fruit flies are one of the most harmful insect species to fruit yields. In AlertTrap, implementation of SSD architecture with different state-of-the-art backbone feature extractors such as Mo-bileNetV1 and MobileNetV2 appear to be potential solutions for the real-time detection problem. SSD-MobileNetV1 and SSD-MobileNetV2 perform well and result in AP@0.5 of 0.957 and 1.0 respectively. YOLOv4-tiny outperforms the SSD family with 1.0 in AP@0.5; however, its through-put velocity is considerably slower, which shows SSD models better candidate for real-time im-plementation. We also tested the models with synthetic test sets simulating expected environmen-tal disturbances. The YOLOv4-tiny had better tolerance to these disturbances than the SSD models.
\end{abstract}

% keywords can be removed

\section{Introduction}
    \hspace{5mm} Agriculture plays an important role in economic growth, and improving crop yield is of great concern \cite{unnevehr2007causes} in Vietnam. On the one hand, insect pesticides can affect the metabolic processes of crops to degrade crop yield and quality \cite{muralidharan2006assessments}. On the other hand, fruit flies are known to cause 50 to 100\% crop loss unless timely interventions are implemented. There are just a small number of fruit fly species have been discovered, namely Bactrocera dorsalis, B. correcta, B. cucurbitae, B. tau, B. latifrons, B. zonata, B. tuberculata, B. moroides and B. albistriga, while some species still remain unidentified. The species which are harmful to fruits are of the common fruit fly species, namely B. cucurbitae and B. tau \cite{vargas2015overview}. In order to optimize crop yields, agricultural workers tend to use a pesticide scheduler rather than take into account the likelihood of pests’ presence in the crop \cite{hazarika2009insect}. Thus, this not only causes a large number of pesticide residues in agricultural com-modities but also brings great pressure to the ecological environment \cite{miller2014sustaining}. The over-use of pesticides is partly because the information about pest species and densities cannot be provided in a timely and accurate way. In contrast, if the information is provided in a timely fashion, it could be possible to take proper prevention steps and adopt suitable pest management strategies including the rational use of pesticides \cite{cho2007automatic} \cite{zhang2009incorporating}.
    
    \hspace{5mm} Traditionally, the information about the environment and pest species is acquired mainly through hand-crafted feature engineering \cite{cho2007automatic} such that workers manually use sensors and compare a pest’s shape, color, texture, and other characteristics with justification from the domain experts. Likewise, counting is typically time-consuming, labor-intensive, and error-prone \cite{zhang2009incorporating}. and therefore, it is urgent and significant to establish an autonomous and accurate pest identification system. There is a growing tendency of utilizing machine vision technology to solve these problems with promising performance in the agriculture research field.
    
    \hspace{5mm} In this work, we focus on developing a solution to detect oriental yellow flies which usually harm citrus fruits such as oranges and grapefruits. We further evaluate the object detection models proposed for yellow fly detection problem in \cite{pham2021alerttrap} by testing the models with test set simulating potential disturbances occurring in real scenario. Additionally, the work presented in this paper will not only focus on the use of different types of object detection algorithms but also apply the TF-LITE format of the models compatible to edge device system such as TPU processors. This direction of study is to develop real-time detection application with the emerging edge computing technology to enhance the performance of the system in terms of detection accuracy, power efficiency, and latency reduction with the purpose of detecting the living fruit flies beside the stuck and dead ones on the trap. Moreover, the article will describe the hardware implementation so that the work can be reproduced and further developed. Our contributions are:
    
    \hspace{10mm} (1)	We constructed, developed, and provided a more in-depth discussion of the end-to-end camera-equipped trap, named AlertTrap with installation of a Lynfield-inspired sticky trap, to instantly detect fruit flies and the solar-energy powering system controlled by a separate Raspberry Pi. This work may also provide more detailed information complementing the hardware description of the system provided in \cite{pham2021alerttrap}.
    
    \hspace{10mm} (2)	Like our previous work in \cite{pham2021alerttrap}, we continue to evaluate three different compact and fast object detection deep learning models, namely SSD-MobileNetV1, SSD-MobileNetV2, and the Yolov4-tiny; nevertheless, we further introduce ar-tificial disturbances imitating inference effects which may compromise the de-tection performance in real-time scenario. Moreover, we also evaluate the SSD-MobileNetV1 and SSD-MobileNetV2 models with their TFLITE format versions on a TPU device. With the results, we compare not only their ability to accurately detect and localize the fruit flies which we had trained them to predict, but also the increase in processing speed as well as the power saving factor.
    
    \hspace{10mm} (3)	Finally, we contributed a dataset of 200 images of fruit flies lying on the sticky trap. The dataset can be used to finetune a pretrained model so that it can be retrained with a short training time and used for the yellow fly detection problem. In addition, we also provide our synthetic data for evaluation with disturbances along with the original data which is disturbance-free. The link to the image source can be found in Data Availability Statement.

\section{Related Works}
\label{sec:relatedW}
    \hspace{5mm} Insect detection techniques can be classified into three system types, namely manual, automatic, or semi-automatic systems. Manual insect detection techniques are known as a process in which trained workers count the trapped flies on a daily basis. These turn out to be error-prone, time-consuming, and labor-intensive, while semi-automatic and automatic systems can address the disadvantages with the replacement of highly accurate and autonomous emerging technological software and hardware.
    
    \hspace{5mm} Specifically, the remaining two types of insect detection systems are often called e-traps as they are fueled by electronic components with extensive computer algorithms such as a center-controlled unit connecting with a camera and the trap actuators. Thus, they are also known as vision-based insect traps. As suggested in the names, the automatic insect detection systems \cite{zhong2018vision,kalamatianos2018dirt,xia2018insect,kaya2014application,doitsidis2017remote,tirelli2011automatic,sun2016automated,philimis2013centralised,moscetti2015feasibility,haff2013automatic,xing2008comparison,potamitis2017automated} are fully autonomous, whereas the semi-automatic ones \cite{sciarretta2019defining,miranda2019developing,shaked2018electronic,wang2017construction} involve human interaction in the loop. For example, in \cite{wang2017construction}, the images of insect body parts are classified in order to aid the human to better categorize the insects. Generally, the e-traps are equipped with a wide range of post-processing techniques to detect and classify trapped insects. These techniques are recognized by the sensor type that is used to capture the existence of insects in the trap. Particularly, they are image-based, spectroscopy-based, and opto-acoustic techniques, which correspond respectively to the visible-light camera, the near-infrared (NIR) camera, and the ultrasound sensor.
    
    \hspace{5mm} The image-based techniques consist of three sub-domain techniques, namely deep learning \cite{zhong2018vision,kalamatianos2018dirt,xia2018insect,kaya2014application} and shallow learning \cite{xia2018insect}, which both are sub-domains in the machine learning field, and image processing \cite{doitsidis2017remote,tirelli2011automatic,sun2016automated,philimis2013centralised,moscetti2015feasibility,haff2013automatic,xing2008comparison,potamitis2017automated} techniques. 
    
    \hspace{5mm} Shallow learning-wise, Kaya et al. \cite{xia2018insect} created a machine-learning-based classifier that can differentiate between 14 butterfly species. The texture and color characteristics are extracted by the writers. A three-layer neural network is used to process the extracted features. The categorization accuracy achieved is 92.85 percent.
    
    \hspace{5mm} The detection approach is based on image processing as described in \cite{tirelli2011automatic,sun2016automated,philimis2013centralised,moscetti2015feasibility}. While image-processing techniques are simpler than deep learning techniques, their accuracy is reasonable (70-80\%) and the system is wired with the illumination environment. How-ever, extensive feature engineering has to take place prior to the classification. Doitsidis and colleagues \cite{kaya2014application} created an image processing method to detect olive fruit flies. By using auto-brightness adjustment, the algorithm first reduces the effect of changing lighting and weather conditions. Then, using a coordinate logic filter improves the edges by amplifying the difference between the dark bug and the bright background. Finally, the technique uses a circular Hough transform followed by a noise reduction filter to identify the trap's limits. The achieved accuracy rate is 75\%. Tirelli et al. \cite{doitsidis2017remote} created a Wireless Sensor Network (WSN) for detecting pests in greenhouses. The image processing technique first removes the effect of light changes from the photos, then denoises them, and finally recognizes the blobs. Sun et al. \cite{tirelli2011automatic} suggested an insect image processing, segmentation, and sorting algorithm for insect "soup" images. In insect "soup" photos, the insects float on the liquid surface. The method was evaluated on 19 soup images by the authors, and it worked well for the majority of them. Using McPhail traps, Philimis et al. created a WSN that detects the olive fruit fly and med-fly in the field \cite{sun2016automated}. WSNs are sensor networks that gather data and may be built to process the information and transfer it to humans. WSNs may also have actuators that respond to specific events. The template comparison algorithm is the detection algorithm. The identification is based on the detection of specific anatomical, patterning, and color characteristics.
    
    \hspace{5mm} Moscetti et al. used Near Infrared Spectroscopy (NIR) to identify infested olives in harvested crops \cite{philimis2013centralised}. The Genetic Algorithm (GA) extracts the features from the collected full spectral data. The retrieved features serve as the input for the classifier. With 93.75 percent accuracy, Quadratic Discriminant Analysis (QDA) is the most accurate. Haff et al. used hyperspectral imaging to identify contaminated mangoes \cite{moscetti2015feasibility}. The algorithm's overall error proportion of high infested samples ranges between 2\% and 6\%, whereas the algorithm's overall error rate for low infested samples is 12.3 percent. In order to detect contaminated cherries, Xing et al. used reflectance and transmittance spectra \cite{haff2013automatic}. Ac-cording to the extent of damage, the cherries were separated into two categories: "acceptable" and "non-acceptable." On transmittance spectra, Canonical Discriminant Analysis (CDA) achieved 85 percent classification accuracy. 
    
    \hspace{5mm} Potamitis et al. \cite{tirelli2011automatic} used opto-acoustic spectrum analysis to construct an olive fruit fly detection system. The opto-acoustic spectrum analysis detects the species of insects based on wingbeat analysis. The authors examined the recorded signal's temporal and frequency domains. The random forest classifier is fed the retrieved features from the time and frequency domains. The random forest classifier had a precision of 0.93, a recall of 0.93, and an F1-Score of 0.93. The opto-acoustic approach, on the other hand, cannot distinguish between different types of fruit flies, including peaches and figs. Furthermore, solar radiation affects sensor readings, and the trap is susceptible to sudden strikes or shocks that cause false alarms on windy days.
    
    \hspace{5mm} Böckmann et al. \cite{oraby2021modeling} utilizes Bag of Visual Words (BoVW) to encode clusters of key points extracted by scale-invariant feature transform (SIFT) into some meaningful local features in a so-called visual codebook. This kind of dictionary is then used to incorporate how frequent each feature appears in each patch of newly extracted key points as the input to train an SVM classifier for different classes of flies as well as one background class for a patch of nothing of interest. In contrast, the precision values decreased after 7 days of the insects remaining on the Yellow Sticky Paper by approximately 20\% compared to the test results of the initialization measurement on day 0. With regard to class mean accuracy, the dictionary size had no obvious influence but on the recall in individual categories. Within the individual categories, the recall of the background class was the highest, as expected. A maximum value of 99.13\% was achieved without differences in color space conversion or dictionary size. The best classification results were achieved with greyscale images and dictionary sizes of 200 and 500 words
    
    \hspace{5mm} Regarding deep learning techniques, Zhong et al. \cite{ding2016automatic} created a deep-learning-based multi-class classifier that can classify and count six different types of flying insects. The You Only Look Once (YOLO) algorithm \cite{redmon2016you} is used for detection and coarse counting. To increase the number of training images required by the YOLO deep learning model, the scientists considered the six species of flying insects as a single class. The authors augment the images with translation, rotation, flipping, scaling, noise addition, and contrast adjustment to extend the data set size. They also employed a pre-trained YOLO to fine-tune its parameters on an insect dataset. Support Vector Machine (SVM) is used for classification and fine counting, with global features. The technique was run on Raspberry PI, with detection and counting performed locally in each trap. The system attained a 92.5 percent average counting accuracy and a 90.18 percent average categorization accuracy. The Dacus Image Recognition Toolkit (DIRT) was created by Kalamatianos et al. \cite{zhong2018vision}. The toolkit includes MATLAB code samples for fast experimentation, as well as a collection of annotated olive fruit fly photos acquired by McPhail traps. On the DIRT dataset, the authors tested various forms of the pre-trained Faster Region Convolutional Neural Networks (Faster-RCNN) deep learning detection technique. Prior to classification, RCNNs are convolutional neural networks containing region proposals that suggest the regions of objects. Faster-RCNN had a mAP of 91.52 percent, where mAP is the average maximum precision for various recall levels. The authors demonstrated that image size has a substantial impact on the detection, but RGB and grayscale images have almost the same detection accuracy. Because Faster RCNN is computationally costly, each e-trap regularly uploads its collected image to a server for processing. Ding et al. created a technique for detecting moth flies \cite{zhang2009incorporating}. Translation, rotation, and flipping are used to enhance the visuals. To balance the average intensities of the red, green, and blue channels, the photos are preprocessed with a color-correcting algorithm. The moths in the photos are then detected using a sliding window Convolutional Neural Network (CNN). CNNs are supervised learning algorithms that use learned weights to apply filters on picture pixels. Backpropagation is used to learn the weights. Finally, Non-Max Suppression is used to remove the overlapping bounding boxes (NMS). Using an end-to-end deep learning neural network, Xia et al. detect 24 kinds of insects in agriculture fields \cite{kalamatianos2018dirt}. A pre-trained VGG-19 network is utilized to retrieve the features. The insect's position is then determined through the Region Proposal Network (RPN). The proposed model had a mAP of 89.22 percent.
    
    \hspace{5mm} Recently, YOLO is once again proving its notable performance in the work \cite{yun2022deep} in pest detection. Especially, the reported results of YOLO v5l by the authors illustrate the mAP of 94.7 percent, where it has the highest recall score of 0.92 among all the other state-of-the-art methods, such as Fast RCNN, Faster RCNN and RetinaNet. The models have been pretrained on COCO dataset \cite{lin2014microsoft} and later fine-tuned on a training dataset of 4480 sub-images made out of 280 images of yellow sticky pheromone traps. However, YOLO v5l is considered slower than YOLOv4. In our previous study, Pham et al. \cite{pham2021alerttrap}, we used Lynfield-inspired trap with naled-and fipronil-intoxicated methyl eugenol \cite{chen2019detection} in replacement of the yellow sticky paper trap combined with object detection system to detect only targeted oriental yellow flies. Unlike the yellow sticky paper, the substance is proved to only attract harmful fruit flies and the detection problem is thus reduced to one-class detection for detecting the existence of the fruit flies and verifying whether the detection is correct. The work showed primary work and provided foundation to further develop real-time system for yellow fly detection in on-field scenario. Compared to \cite{yun2022deep}, the application Single Shot Multibox Detector with variant backbones and YOLOv4-tiny show significant speed performance to YOLO v5l, while taking the raw images as input instead of segmented sub-images. Nevertheless, the work also showed limitation of ap-plying detection models on edge device due to the slow processing speed, which will be further addressed in this concurrent work.

\section{Methodology}
\label{sec:method}
\subsection{Overview of the Trap System}
\label{sec:overview}
    \hspace{5mm} Most of the time, insects are not stationary, so it is difficult to get a clear image of flying insects. In studies \cite{wang2012identification},\cite{wang2012new},\cite{kang2014identification},\cite{kang2012identification}, the authors chose insect specimens that were well-preserved in an ideal laboratory environment to capture images of the insects at high resolution. However, since fewer environmental factors are considered in this method, it is limited in specific applications. In this study, we designed a unique automatic au-tonomous environment data reading and pest identification system to try to eliminate the above problems.
    
    \hspace{5mm} Being largely motivated by preventing the oriental fruit flies from destroying citrus fruits such as oranges and grapefruits, we come up with a trap which targets only that one type of the species, which is specifically named as B. Dorsalis. This can be achieved by replacing the yellow sticky paper with the naled-and fipronil-intoxicated methyl eugenol attractant to assure only B. Dorsalis flies are lured into the trap. It eases the classification and counting process as no other insects will get attracted by the methyl eugenol attractant \cite{chen2019detection}. The objects can be further reassured by the object detection system before getting counted.
    
    \hspace{5mm} The system involves a two-fold setting: a) an electronic system reads environment data with a sticky trap installed and a digital camera is set up to collect images of the flies, b) the object detection software to recognize fruit flies on the image before sending all information (environmental data and number of fruit flies) via email or SMS to alert farmers independently. The whole system is autonomous and powered by a solar system. This system is implemented on an Arduino Uno and Raspberry Pi system. The results provide precise prevention and treatment methods based on the combination of pest information and other environmental information. Based on this edge computing design, the computation pressure on the server is alleviated and the network burden is largely reduced.
    
    \hspace{5mm} The edge-computing traps are designed to work separately and individually report the count of fruit flies to the farmers. They are spread, based on the effectiveness of the attractant, such that each 2-3 devices can cover the area of 1000 square meters.

\subsection{Hardware}
\label{sec:hardware}
\begin{figure}
	    \centering
	        \includegraphics{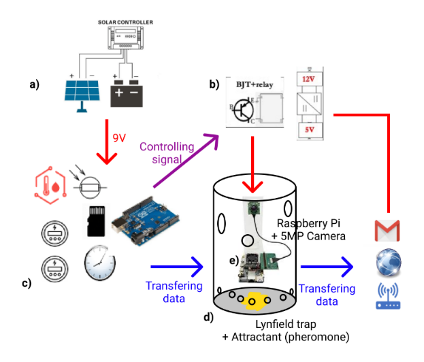}
	    \caption{Overview of the trap system consisting of a) The solar panel system, b) The operation system, c) The sensor system, d) The modified Lynfield trap and e) The object detection system \cite{pham2021alerttrap}.}
	    \label{fig:fig1}
    \end{figure}

    \hspace{5mm} Overall, the hardware part of the system consists of five interconnected subsystems with distinctive functions and behaviors, which are described in Figure \ref{fig:fig1}, which is also shown in \cite{pham2021alerttrap}, namely the solar panel system, the control system, the sensor system, the trap, and the object detection and communication system.
    
    \hspace{5mm} The power system of the trap contains of a solar panel, a battery, and a solar charge controller (Figure \ref{fig:fig1}a). The solar panel converts the solar energy to DC current with 830 mA to power the trap system. The converted energy is stored in an electrochemical energy storage with a capacity of 5 Ah and a voltage of 12 V. The Arduino in the operation system will check voltage of the battery with a voltage sensor to make sure the battery voltage above a certain level required for the system’s operation. If the condition is not met, the object detection module will not be operated. The Pulse Width Modulation (PWM) solar charge controller is used in order to control the device voltage, open the circuit, and halt the charging process if the battery voltage is above a certain level. 
    
    \begin{figure}
	    \centering
	        \includegraphics{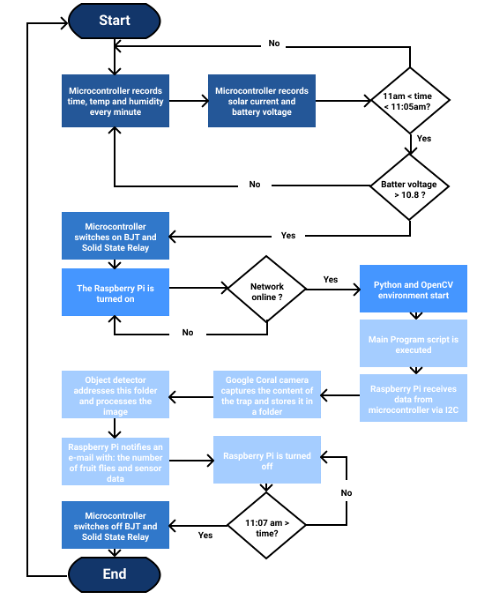}
	    \caption{Flow chart of the trap system.}
	    \label{fig:fig2}
    \end{figure}
    
    \hspace{5mm} The operation system (Figure \ref{fig:fig1}b) is controlled by an Arduino microcontroller board. As aforementioned, the Arduino module reads the battery voltage with a voltage sensor from the sensor system to decide whether to turn on or off the object detection system, which is controlled by the Raspberry module. The SSR10D is used to control activate and deactivate the object detection system. The SSR10D is a solid-state relay and uses lower power electrical signal to generate an optical semiconductor signal as an activate signal for the opto-transistor to allow high voltage going into and powering the device’s output device, which is the Raspberry device in this case. In addition, the lower electrical signal is the output from the 2N2222 bipolar junction transistor receiving control signal from the Arduino module. Hence, the Arduino can stop the Raspberry Pi 3b+ computer drawing current from the solar system after it is shut down. The sensor system (Figure \ref{fig:fig1}c) takes responsibility for measuring the three important factors, temperature, humidity and light. Also, it records the current created by the solar system and the voltage battery. The humidity and temperature, which also affect the living environment of the yellow flies, are measured with the AM2315 I2C sensor. RGB and clear light is measured with the TCS34725 light sensor with IR filter and white LED. In addition to sensor system, INA219 is used to read the solar current and battery voltage information. Moreover, a DS1307, which is a battery-backed real time clock (RTC), is used to help the microcontroller keep track of time. These information from the sensors along with their corresponding time are stored in a SD card attached on the device. These two factors, the operation system and sensor system, help the microcontroller decide whether to turn on the object detection or not. The object detection system, shown in Figure \ref{fig:fig1}e, is operated by the Raspberry Pi 3b+ and collect images for its fruit fly detection algorithm with a waveshare Pi camera with 5 MP. The camera is placed at the top of a double-size Lynfield shape trap with several holes at the bottom, shown in Figure \ref{fig:fig1}d. To attract and capture only the yellow flies, methyl eugenol is used as the attractant to the insects \cite{chen2019detection}, which later help to simplify the de-tection and classification problem. The Raspberry Pi module will receive data from the sensor system and sends all data to the notification system to notify or alert farmers about the environment data and the number of detected fruit flies through email or SMS. The behavior of the whole system is described in the flow chart shown in Figure \ref{fig:fig2}.

\subsection{Software - Object detection pipelines}
\label{sec:software}
    \hspace{5mm} The architectures used to train the yellow fly detection models are SSD with MobilenetV1 and MobilenetV2 backbones, and YOLOv4-tiny. The selected models are all single-stage detection models since, compared to their counterpart, the two-stage detection models, the single-stage detection models have been shown to have a faster processing speed with a competitive performance. The pre-trained models of these architectures were fine-tuned with our proposed insect dataset so that they can be used for the yellow fly detection application, as fine-tuning is also one of the common solutions for data scarcity problem in object detection. Because the models had been trained with COCO dataset \cite{lin2014microsoft}, which is a large dataset having over 200,000 labeled images with 1.5 million object in-stances for 80 object categories, and, hence, contains common features for object detection problem, fine-tuning the models with 200 yellow fly images helped the models perform the yellow fly detection task.
    
\subsubsection{SSD}
\label{sec:ssd}
    \hspace{5mm} To solve the real-time object detection in the yellow fly detection problem, variants of Single-Shot Multibox Detector (SSD) are used. The SSD method was first proposed in \cite{liu2016ssd} by Wei Liu et al. and described as a one-stage object detection method that completely omits the region proposal and pixel/feature re-sampling stages used in region proposal-based techniques such as Faster-RCNN. The SSD network is based on a feed-forward network that uses default bounding boxes with different shapes, ratios, and scales in order to produce a fixed-size collection of bounding boxes with corresponding shape offsets and confidence scores \cite{liu2016ssd}. In addition, the early layers of the network are based on a standard image classification without classification layers, which is called the base network \cite{liu2016ssd}. In this work, MobileNetV1 and MobileNetV2 are used as base networks for the SSD detection models. The elimination of region proposal and pixel/feature re-sampling stages helps to improve the processing speed of the model compared to two-stage techniques such as Faster-RCNN with a small trade-off in the model’s accuracy, which enables the implementation of real-time object detection with high accuracy on embedded system for yellow fly detection problem.
    
\subsubsection{MobileNetV1}
\label{sec:mobilenetv1}
    \hspace{5mm} The approach was first proposed in \cite{howard2017mobilenets} by A. Howard et al. and was described as a lightweight deep neural network for mobile and embedded system applications with an efficient trade-off between latency and accuracy. The model is based on depth wise separable convolution including depth wise convolution layer which is used to apply a single filter per input channel, and point-wise convolution layer, which creates a linear combination of the output of the depth wise layer. In addition, to construct the model further less computationally expensive, width multiplier, which is used to thin the network uniformly at each layer, and resolution multiplier, which is applied to input images and the internal representation of each layer, were introduced as a hyper-parameter to tune and choose the size of the model.

\subsubsection{MobileNetV2}
\label{sec:mobilenetv2}
    \hspace{5mm} The MobileNetV2 approach was first presented in \cite{sandler2018mobilenetv2} by M. Sandler et al. The approach is built based on the MobileNetV1; therefore, it also makes use of the depth wise separable convolution architecture which consists of depth wise convolution layer and 1x1 point-wise convolution layer. In addition, the approach also utilizes linear bottleneck layers in convolutional blocks to optimize the neural architecture \cite{sandler2018mobilenetv2}. Moreover, inverted residual design is also used in the model to implement shortcuts between bottlenecks with the purpose of improving the ability of gradient propagation across the multiplier layers. Nevertheless, the implementation of the inverted design also showed better performance and significantly more memory efficiency in the work \cite{sandler2018mobilenetv2}.

    \hspace{5mm} The training and evaluation of the SSD with MobileNetV1 and MobileNetV2 base networks is based on the pre-trained models provided in the Object Detection API in TensorFlow Model Garden \cite{huang2017speed}. The models were also trained on the Google Colab Pro environment to utilize the provided GPUs for the training purpose.

\subsubsection{YOLOv4-tiny}
\label{sec:YOLOv4-tiny}
    \hspace{5mm} The YOLOv4 model is a method for detecting objects that was developed from the YOLOv3 model proposed in \cite{redmon2018yolov3}. The YOLOv4 approach was created by Alexey Bochkovskiy, Chien-Yao Wang, and Hong-Yuan Mark Liao \cite{bochkovskiy2020yolov4}. It is twice as fast as EfficientDet in terms of performance. Furthermore, when compared to YOLOv3, AP (Average Precision) and FPS (Frames Per Second) in YOLOv4 have increased by 10\% and 12\%, respectively. This is because a CSPDarknet53 backbone and a PANet path-aggregation neck along with the YOLOv3 head make up for the YOLOv4 architecture. YOLOv4-tiny \cite{jiang2020real} is the compressed version of YOLOv4. Based on YOLOv4, it is suggested that the network topology should be simplified, and parameters should be reduced so that it may be implemented on mobile and embedded devices. The YOLOv4-tiny model can be trained in a shorter time than the YOLOv4.

\section{Evaluation}
\label{sec:evaluation}
    \hspace{5mm} The proposed models, YOLOv4-tiny, SSD-MobileNetV1, and SSD-MobileNetV2, were evaluated by using performance metrics such as precision, recall, F1 score, mean IoU, and average precision (or AP). In addition, the models’ processing time was also evaluated to check for real-time application feasibility. The processing time of the models was calculated while they were run on CPU, GPU, and TPU hardware. The hardware provided by Colab Pro service was Tesla V100-SXM2-16GB, TPU V2, and Intel(R) Xeon(R) CPU @ 2.30GHz, which were used to find the average processing time of the models on GPU, TPU, and CPU respectively. Moreover, besides the regular test set, the models were also checked with synthetic test images originating from the original test set. The aug-mentation effects were implemented to simulate disturbances that can be captured in practice, such as leaf, insect, dust, and flaring effect.

\subsection{Training Dataset}
\label{sec:trainingset}
    \hspace{5mm} The contributed 200 images of fruit flies in the trap are consolidated into two parts, namely training dataset and test dataset, with the proportion of 75\% and 25\%, respectively. Therefore, there are 150 images being used to train the models incorporated in this paper. The remaining 50 images are responsible for evaluating the performance of such models.

\subsection{Test Dataset}
\label{sec:testset}
\begin{figure}
	    \centering
	        \includegraphics{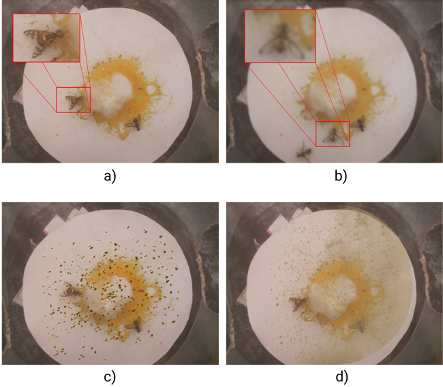}
	    \caption{An example of an original image and its synthetic versions with a) original image, b) blurry filter c) adding salt-pepper disturbance, and d) adding dust disturbance.}
	    \label{fig:fig3}
    \end{figure}

    \hspace{5mm} For the evaluation of the proposed models, the original test set, which includes 50 images, was used along with other four synthetic datasets generated from the original test images. The synthetic datasets were used to simulate the common disturbances which could be captured in real-life scenarios such as blurry, dust, salt-pepper and flaring effect. Therefore, in general, the total test set has 250 images, in which 50 images come from the original dataset and the other 150 images are copies of them with augmentation effects. The example of an original image and its synthetic versions are shown in Figure \ref{fig:fig3}.
    
    \hspace{5mm} The synthetic images with dust and salt-pepper augmentation are to simulate the disturbances coming from the weather and environmental conditions, while the blurring augmentation is used to simulate foggy or out-of-focus issues captured by the camera in the field and maybe disrupt the classification of the targeted object class.
    
\subsection{Evaluation Metrics}
\label{sec:metrics}
    \hspace{5mm} The performance of the trained detectors was evaluated by being tested with a test dataset. The true positive (TP), false positive (FP), and false negative (FN) were counted from the detection results and used to find the precision, recall, and F1 score metrics. TP is the number of correctly detected objects, while FP shows the amount of the falsely detected object and FN informs the number of targeted objects which were missed during the detection process. The three metrics can then be used to evaluate further aspects of the models’ performance such as precision, recall, F1-Score, and average precision, which were also used for model evaluation in this work. In addition, the average processing time of the models during the detection process was also measured and used as another evaluation criterion, and Mean IoU was also used to evaluate the localization of the detection.
    
\subsubsection{Precision}
\label{sec:precision}
    \hspace{5mm} Precision is an evaluation metric calculated with true positives and false positives. The metric indicates how accurate the detector’s performance is by showing the percentage of correctly detected objects out of the total detected objects. The relationship between the precision, true positive and false positive numbers can be expressed as the following:
    \begin{equation}
        Precision=\frac{TP}{TP+FP}=\frac{TP}{All Detections}
    \end{equation}
    
\subsubsection{Recall}
\label{sec:recall}
    \hspace{5mm} Recall is an evaluation metric calculated with counted true positives and false neg-atives. The recall value decreases when the number of false negatives is increased. The metric measures how many objects are missed by the evaluated detector by showing the percentage of correctly detected objects out of the total true objects. The relationship between the recall and true positive and false negative numbers can be expressed as the following:
    \begin{equation}
        Recall=\frac{TP}{TP+FN}=\frac{TP}{All Groundtruths}
    \end{equation}

\subsubsection{F1-Score}
\label{sec:f1}
    \hspace{5mm} F1-Score is an evaluation metric calculated with both precision and recall metrics, or with all counted true positives, false positives, and false negatives. The F1-Score metric combines both precision and recall with equal weight to show the balance and relative relation between the precision and recall metrics. The mathematical expression of the F1-Score metric regarding other evaluation metrics is shown as:
    \begin{equation}
        F1=2*\frac{Precision*Recall}{Precision+Recall}
    \end{equation}
    
\subsubsection{Average Precision}
\label{sec:AP}
    \hspace{5mm} By adjusting the confidence score threshold, the precision and recall will also be changed accordingly. By calculating the mean value of the precision score corresponding to the confidence score threshold varied from 0 to 1, average precision, or AP for short, can be found.
    
\subsubsection{Mean IoU}
\label{sec:mIOU}
\begin{figure}
	    \centering
	        \includegraphics{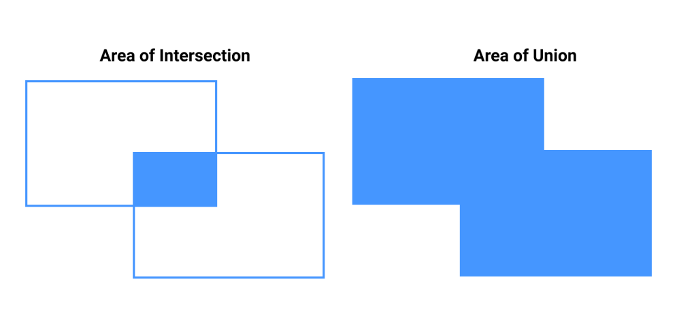}
	    \caption{Area of Intersection (left) and Area of Union (right) examples.}
	    \label{fig:fig4}
    \end{figure}

    \hspace{5mm} IoU stands for intersection over union. The intersection can be understood as the over-lapping area between a ground-truth bounding box and its corresponding detected bounding box, while the union is the total area of the ground-truth bounding box and the corresponding detected bounding box with the omission of the overlapping area. The graphical representation of the mentioned definition is shown in Figure \ref{fig:fig4}.
    
    \hspace{5mm} Then, the IoU is the ratio of the intersection area over the area of union of the two bounding boxes. The higher the value of the IoU, the more matched the two bounding boxes are. Therefore, the mean IoU, which is the mean value of the IoU values of all available pairs of detected bounding boxes and their corresponding ground-truth bounding boxes, shows the localization of the examined algorithm. The mathematical relation between the IoU, the area of intersection, and the area of union can be expressed with the following formula:
    \begin{equation}
        IoU=\frac{Area of Intersection}{Area of Union}
    \end{equation}
    
\subsubsection{Processing Time}
\label{sec:processingtime}
    \hspace{5mm} To examine the feasibility of real-time implementation of the trained models, the models’ processing time was also measured and used as an evaluation metric. The processing speed of a detector is found by calculating the mean processing speed of the detector during its detection operation over 250 test images.

\subsection{Experimental Results}
\label{sec:exres}
\subsubsection{Precision, Recall, F1-score, mean IoU, and AP evaluation}
\label{sec:nottimeresult}
    \hspace{5mm} In this work, the models were evaluated for F1-score, mean IoU, and AP evaluation metrics with different choices of IoU threshold value. The following shows the perfor-mance tables of the trained detectors through different testing scenarios with IoU threshold values of 0.25, 0.50, and 0.75, which may reflect the performance of the algo-rithms with different localization demands. Additionally, in this paper, the result per-formance of the models in normal testset or disturbance-free is further discussed from the discussion in \cite{pham2021alerttrap}. The discussion presented here will complement the evaluation shown in \cite{pham2021alerttrap}.

    \hspace{5mm} \textbf{a) Normal Testset}

\begin{figure}
	    \centering
	        \includegraphics[width=0.65\textwidth]{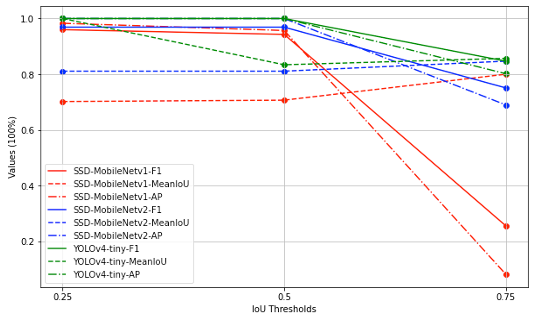}
	    \caption{Evaluation results on Normal testset.}
	    \label{fig:fig5}
    \end{figure}

    \hspace{5mm} In the normal test case with no augmented and synthetic disturbances, it can be observed in all IoU thresholds that YOLOv4-tiny is the model having the best performance in all aspects among the three examined algorithms, YOLOv4-tiny, SSD-MobileNetV1, and SSD-MobileNetV2. Especially, for IoU threshold values 0.25 and 0.5, the YOLOv4-tiny model achieved perfect F1-Score, and AP metrics with the value of 1.0. Moreover, with high localization constraints, YOLOv4-tiny still has good F1-score which is 0.847. In addition, the detection from the YOLOv4-tiny model also has great localization based on the mean IoU metric, which is 0.834 for IoU threshold 0.25 and 0.50, and 0.857 for IoU threshold 0.75. Nevertheless, SSD-MobileNetV2 also has comparable performance to the performance of YOLOv4-tiny, since for IoU threshold 0.25 and 0.5, SSD-MobileNetV2 also achieved a great F1-Score, and AP metrics with the values of 0.969 and 1.0 respectively. In the extreme case of IoU threshold 0.75, SSD-MobileNetV2 can also have good performance with values of 0.751 for F1-score, and 0.69 for AP metric. In addition, SSD-MobileNetV2’s detection also has similar localization compared to YOLOv4-tiny, which are 0.811 for IoU threshold 0.25 and 0.50 and 0.847 for IoU threshold with the value of 0.75. The SSD-MobileNetV1 also has good performance for IoU threshold 0.25 and 0.50; however, with IoU threshold of 0.75, the model’s performance was heavily compromised. More-over, in all three IoU threshold cases, the SSD-MobileNetV2 has better performance than SSD-MobileNetV1 in all aspects. The performance results on different localization constraints are shown in Figure \ref{fig:fig5}. The results can also be found in \cite{pham2021alerttrap}.
    
    \hspace{5mm} \textbf{b) Blurry Testset}

\begin{figure}
	    \centering
	        \includegraphics[width=0.65\textwidth]{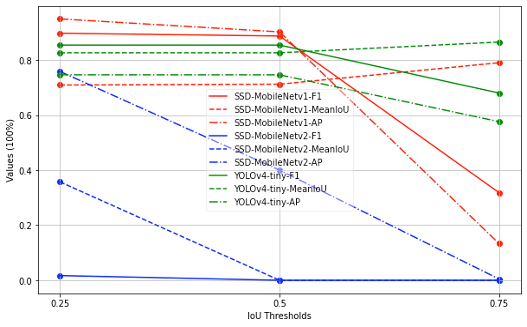}
	    \caption{Evaluation results on Blurry testset.}
	    \label{fig:fig6}
    \end{figure}

    \hspace{5mm} In Blurry testset, a Box filter of size 30x30 is convolved over the images in the Normal testset to create the blur effect which aims to replicate the foggy weather inside the trap or the out-of-focus issue of the camera. Generally, this testset changes the overall features of the fruit flies because of the averaging effect of the filter; therefore, the models might not work as expected, which is clearly shown via the results of SSD-MobileNetV2. However, According to Figure \ref{fig:fig6}, YOLOv4-tiny still performs stably on the testset. Specifically, YOLOv4-tiny can maintain its metrics values in the variation range of 0.2 throughout three IoU thresholds; whereas, SSD-MobileNetV2 is totally collapsed when the IoU threshold increases and SSD-MobileNetV1 significantly drops in F1-Score and AP by a value of 0.6 at the extreme case IoU threshold.
    
    \hspace{5mm} \textbf{c) Salt-pepper Testset}

\begin{figure}
	    \centering
	        \includegraphics[width=0.65\textwidth]{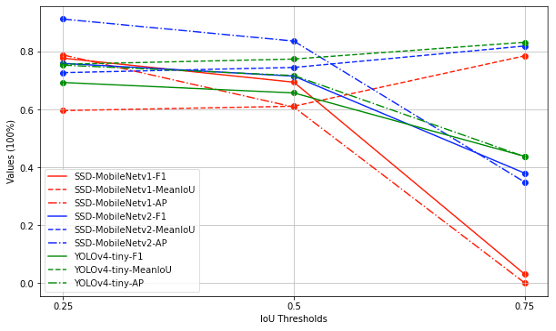}
	    \caption{Evaluation results on Salt-pepper testset.}
	    \label{fig:fig7}
    \end{figure}

    \hspace{5mm} The salt-pepper disturbances imply the appearance of unwanted tiny fraction of leaves or other insects that are accidentally flown into the trap by the wind. It is named “Salt-pepper” because the effect looks visually like salt and pepper on a dish, which is not related to the salt-and-pepper noise in image processing point of view. It can be observed from Figure \ref{fig:fig7} that YOLOv4-tiny, with extreme localization constraint, IoU threshold 0.75, the model outperformed other models, and it also has best mean IoU in all IoU threshold constraints. However, adding the Salt-pepper disturbance makes all three models’ performance worse than their performance on the remaining testsets. Specifically, SSD-MobileNetV1 suffers the most where in the extreme case, both its F1-Score and AP drop to 0.
    
    \hspace{5mm} \textbf{d) Dust Testset}

    \hspace{5mm} With test images with added dust disturbance, it can be observed in all IoU thresholds that YOLOv4-tiny has the best performance in all aspects. Nevertheless, compared to its performance with the normal dataset, it can be observed that the disturbance effect did compromise the performance of the model based on the F1-score. The same conclusion can also be drawn for SSD-MobileNetV1 and SSD-MobileNetV2 models based on their performance on the testset. In addition, the dust disturbance also proves to have a great negative effect on the SSD-MobileNetV2 model when the extreme localization constraint is applied, since while in the Salt-pepper testset with IoU threshold of 0.75, the SSD-MobileNetV2 could still perform well with an F1-Score value of 0.38, while with dust disturbance, the model can only achieve 0.251 in F1-Score. Based on the mean IoU metric, the dust disturbance also decreases the localization of all three models’ detection results.
    
    \begin{figure}
	    \centering
	        \includegraphics[width=0.65\textwidth]{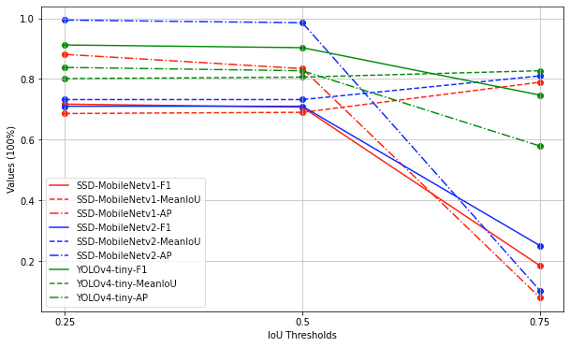}
	    \caption{Evaluation results on Dust testset.}
	    \label{fig:fig8}
    \end{figure}
    
\subsubsection{Processing time evaluation}
\label{sec:timeresult}
    \hspace{5mm} In this work, the model candidates are examined with CPU, GPU, and TPU hardware. Moreover, the SSD-MobileNetV1 and SSD-MobileNetV2 models with TFLITE convert support were also converted into TFLITE format which is compatible with TPU and can exploit the advantage of the hardware. Table 1 shows the processing time of each model on CPU, GPU, and TPU hardware. It can be observed that SSD models have a faster processing time than the YOLOv4-tiny model on GPU, CPU, or TPU hardware. Moreover, the TFLITE version of SSD-MobileNetV1 and SSD-MobileNetV2 models run on TPU is much faster than SSD-MobileNetV1, SSD-MobileNetV2, and YOLOv4-tiny in-ference graphs run on the same device. In addition, compared to the work in \cite{pham2021alerttrap}, we update the processing speed in FPS for TFLITE models with SSD-MobileNetV1 and SSD-MobileNetV2 architecture on a TPU device. TPU processors can be found in edge device such as Google Coral Dev board, and by comparing the processing speed of the models in TFLITE format in TPU shown in this work and the processing speed of the models in Raspberry Pi shown in \cite{pham2021alerttrap}, it can be shown that with the same architectures, the TFLITE models’ processing speed on a TPU device is approximately 6 times faster than the inference models on Raspberry Pi. This shows that TFLITE model on TPU edge device would be a more feasible for real-time application implementation. 
    
    \begin{center}
        \begin{tabular}{||c c c c||} 
             \hline
             Models & \makecell{\textbf{CPU} \\ Intel(R) Xeon(R) CPU @ 2.30GHz} & \makecell{\textbf{GPU} \\ Tesla V100-SXM2-16GB} & \makecell{\textbf{TPU} \\ Google TPU v2} \\ [0.5ex] 
             \hline\hline
             SSD-MobileNetV1 & 3.414 FPS & 29.140 FPS & 1.025 FPS \\ 
             \hline
             SSD-MobileNetV2 & 3.485 FPS & 21.000 FPS & 1.024 FPS \\
             \hline
             SSD-MobileNetV1 TFLITE & -- & -- & 8.743 FPS \\
             \hline
             SSD-MobileNetV2 TFLITE & -- & -- & 10.058 FPS \\
             \hline
             YOLOv4-tiny & 1.282 FPS & 1.207 FPS & 0.545 FPS \\ [1ex] 
             \hline
        \end{tabular}
    \end{center}

\section{Discussion}
\label{sec:discussion}
    \hspace{5mm} The assessments in this research are dedicated to search for the most appropriate object detection method among the current state-of-the-art algorithms which have been implemented for insect and fly recognition under our hardware constraints and problem definition. As we only target one type of fruit flies that particularly causes harm to the citrus fruits, we have replaced the yellow sticky paper with a white disc containing the special attractant as a hard refinement to pick up only the flies we are interested in. The object detection problem is then simplified to only one-class object detection, which eases the need of exhausting feature extraction. However, the general constraints, such as correctness and fastness, for an object detection task on an edge-device still hold since early detection and separation of the infected areas are extremely important to the fruit yield. Ultimately, SSD-MobileNetV1, SSD-MobileNetV2 and YOLOv4-tiny are the best can-didates for these requirements because they utilize extracted features from a backbone classification model to automatically propose object-related regions instead of using a region-proposal module to pool the related regions before classifying them as many two-stage object detection models, such as Fast-RCNN and Faster-RCNN,  normally do.
    
    \hspace{5mm} Regarding the correctness, YOLOv4-tiny clearly outperforms the two SSD models over all of the evaluations on four different types of testset with very high and stable results. This could make YOLOv4-tiny become the most probable candidate, because YOLOv4-tiny demonstrates a robust testing performance towards citrus fruit flies detec-tion although it has been fine-tuned only on a training dataset without augmentation effects. SSD-MobileNetV2 shows appropriate robustness given its small number of trainable parameters by yielding good results in two over four testsets, while SSD-MobileNetV1 only works with the original testset. Nevertheless, SSD-MobileNetv2 fails dramatically with the Blurry testset, which simulates a very frequent event that could happen in a fruit field. By all means, YOLOv4-tiny is no doubt the chosen one among the three methods if we would not have taken other aspects into account.
    
    \hspace{5mm} Conventionally, highly accurate object detection methods trade their processing speed for its better performance due to the employment of more parameters in their architecture. YOLOv4-tiny is not an exception where its processing speed is far from real-time (1.2 FPS compared to 30 FPS). While missing a fraction of time could lead to undetectable event in which the flies appear, our second choice which is the SSD-MobileNetV2 model should be taken into account. In order to realize this choice after extensive performance analysis with four different testsets, SSD-MobileNetV2 must have been fine-tuned with more augmented versions of the original training dataset before going to production to leverage its robustness to the level of YOLOv4-tiny while retaining its processing speed. Moreover, TFLITE version of SSD-models are also tested on a cloud TPU Google engine, TPUv2, for the feasibility of edge-device deployment.

\section{Conclusion and Outlook}
\label{sec:conclusion}
    \hspace{5mm} Experimental results show that the Raspberry Pi system successfully gained envi-ronmental data and number of counted pests which were transferred to email addresses through the 4G network. The full YOLO version cannot run real time on Raspberry Pi which poses the need of a lighter Object detection algorithm for future research. 
    
    \hspace{5mm} From the results, it can also be concluded that in general, YOLOv4-tiny has the best performance among the three model candidates. Nevertheless, SSD-MobileNetV2 also has a comparable performance to YOLOv4-tiny. Moreover, the SSD-MobileNetV2 model also outperforms the YOLOv4-tiny model in some test scenarios with synthetic disturbances. In addition, SSD models have a clear advantage over YOLOv4-tiny in processing time criterion, which makes real-time detection application with high accuracy feasible. Furthermore, the TFLITE versions of SSD models also process faster than the SSD models’ inference graph on TPU hardware, suggesting a feasibility of real-time implementation of the SSD models on edge devices with TPU processors such as Google Coral Dev Board with edge TPU. Therefore, in the future work, Google Coral Dev Board will be im-plemented on the system, which can be used to compare with the Raspberry Pi 3b+’s in accuracy and processing time aspects. In addition, infield operation of the system will be tested to check the system’s practicability and for further improvement. In addition, from the test result with the synthetic test sets, it can be seen that the SSD family models were susceptible to disturbance and noise compared to the YOLOv4-tiny model. Therefore, our next attempt is to improve the SSD models’ performance training the detectors with augmented and synthetic data synthesized from the original dataset. Moreover, by building several trap devices, we will try to apply federated learning on the multiple on-field traps so that the detection algorithm can be trained, and improved while being applied on the field. Hence, the detection performance of the traps can be further boosted. Furthermore, we also would like to further develop our detection solution to other types of insects so that it may not only enhance the yellow fly detection performance but also make the solution applicable for other insect detection problems. To achieve the goal, we will need to expand our dataset so that it would contain other types of insects.

\section{Data Availability Statement}
 \hspace{5mm} The image dataset can be found at the following link: https://github.com/a11to1n3/AlertTrap-Dataset.

\bibliographystyle{unsrtnat}
\bibliography{references}

\end{document}